# SOME EXTENSIONS OF PROBABILISTIC LOGIC

Su-shing Chen
Department of Computer Science
University of North Carolina, Charlotte
Charlotte, NC 28223

## ABSTRACT

First order logic is a very useful tool for knowledge representation, inference engine and AI programming. Recently, Nilsson has proposed the probabilistic logic and a probabilistic inference scheme. We extend it to evidential logic in the Dempster-Shafer fashion. In terms of multi-dimensional random variables, we observe that the permissible probabilistic interpretation vector of Nilsson consists of the joint probabilities of logical propositions. Thus, probabilistic logic is subsumed by probability theory.

## 1. INTRODUCTION

In [1], Nilsson proposed the probabilistic logic in which the truth values of logical propositions are probability values between 0 and 1. It is applicable to any logical system for which the consistency of a finite set of propositions can be established. The probabilistic inference scheme reduces to the ordinary logical inference when the probabilities of all propositions are either 0 or 1. This logic has the same limitations of other probabilistic reasoning systems of the Bayesian approach. Moreover, for AI applications, Nilsson's consistency is not a very natural assumption. We have some well known examples: {Dick is a Quaker, Quakers are pacifists, Republicans are not pacifists, Dick is a Republican} and {Tweety is a bird, birds can fly, Tweety is a penguin}.

In this paper, we shall consider the space of all interpretations, consistent or not. In terms of frames of discernment, the basic probability assignment (bpa) and belief function can be defined. Dempster's combination rule is applicable. This extension of probabilistic logic may be called the evidential logic. For each proposition $s$, its belief function is represented by an interval $[Spt(P), Pls(P)]$. When all such intervals collapse to single points, the evidential logic reduces to

43

probabilistic logic (in the generalized version of not necessarily consistent interpretations). Of course, we get Nilsson's probabilistic logic by further restricting to consistent interpretations.

We shall give a probabilistic interpretation of probabilistic logic in terms of multi-dimensional random variables. Let us consider a finite set $S = \{s_1,..., s_n\}$ of logical propositions. Each proposition $s_i$ may have true or false values and may be considered as a random variable. We have a probability distribution for each proposition. The n-dimensional random variable $(s_1,..., s_n)$ may take values in the space of all interpretations of $2^n$ binary vectors. We may compute absolute (marginal), conditional and joint probability distributions. It turns out that the permissible probabilistic interpretation vector of Nilsson [1] consists of the joint probabilities of S. Inconsistent interpretations will not appear, because they are zeros. By summing appropriate joint probabilities, we get probabilities of individual propositions or subsets of propositions. Since the Bayes formula and other techniques are valid for n-dimensional random variables, the probabilistic logic is actually subsumed by probability theory.

## 2. EVIDENTIAL LOGIC

We consider a set $S = \{s_1,..., s_n\}$ of logical propositions. In [1], Nilsson introduced the concept of a binary semantic tree of S which represents $2^n$ possible interpretations. At each node, we branch left or right, indicating a proposition and its negation. Below the root, we branch on $s_1$, then on $s_2$, and so on. We may assign truth values to these branches according to propositions and their negations. Since we are dealing with evidential reasoning, logically inconsistent interpretations are kept.

Let $\Theta$ denote the set of $2^n$ interpretations. Each proposition $s_i$ in S is either true or false in $\Theta$, thus, is completely characterized by the subset $A_i$ of $\Theta$ containing those interpretations wher $s_i$ is true. We shall say that $s_i$ is supported by $A_i$. A basic probability assignment (bpa) m is a generalization of a probability mass distribution and assigns a number in $[0,1]$ to every subset $A$ of $\Theta$ such that the numbers sum to one.

$$m{:}\Theta \rightarrow [0,1], \sum_A m(A) = 1.$$

If a proposition $s_i$ is supported on a subset $A_i$ of $\Theta$ , then for all $A \subseteq A_i$

$$Spt(s_i) = \sum_{A} m(A), \, Pls(s_i) = 1 - Spt(\neg s_i).$$

We have a system of equations of $Spt(s_i)$ and $Pls(s_i)$ for i = 1,...,n which gives the collection $\{[Spt(s_i), Pls(s_i)]\}$ of evidential intervals. If the evidential interval of a proposition s collapses to a point, then we have the probability (or generalized truth value) of s [1]. We may very well have a mixed situation which allows some intervals and some points for the set S of logical propositions.

There are various kinds of evidential reasoning schemes. One may try to extend the update scheme of Pearl in [11]. Nilsson's probabilistic inference [1] can be extended as follows:

1.   Construct S′ by appending a new proposition s to S.

2.   Extend the binary semantic tree.

3.   Construct $\Theta'$.

4.   Find a bpa m′ from the system of evidential intervals of S.

5.   Calculate the evidential interval of s using m′.

The Dempster combination rule applies also. For two bpa $m_1$ and $m_2$, the combination $m_1 + m_2$ may be calculated over intersecting subsets of $\Theta$. Here, the binary semantic tree seems to be nice for dealing with the hierarchical hypothesis space [10].

## 3. SEMANTICS AS RANDOM VARIABLES

Given the set S = $\{s_1,...,s_n\}$ of n logical propositions, the binary semantic tree gives $2^n$ possible assignments of truth values {0,1} to S. For each $s_i$ , there are two possible outcomes 0 and 1. A probability measure on the Borel field on {0,1} gives a probability distribution p of the random variable $s_i$. The values of p at two outcomes 0 and 1 are nonnegative and they sum to be one.

45

Without much ambiguity, we shall use the same symbol S for the n-dimensional random variable

$$S = (s_1, ..., s_n).$$

S may assume value in the set of n-dimensional binary vectors $v_j$, where $j = 1, ..., 2^n$, which are the interpretation vectors in [1]. The permissible probabilistic interpretation vector of [1] turns out to be the joint probability distributions $p(v_j)$, $j = 1, ..., 2^n$. For inconsistent $v_j$, its joint probability $p(v_j)$ is zero and does not contribute to the permissible probabilistic interpretation vector.

In [1], a consistent probabilistic valuation vector over S is computed by multiplying the sentence matrix M to a permissible probabilistic interpretation vector. The components of this valuation vector are the generalized truth values or probabilities of propositions $s_i$ in $\overset{.}{S}$. These components are the marginal (or absolute) probability distributions $p(s_i = 1)$ for $i = 1, ..., n$, because the marginal probability distribution of any component of a n-dimensional random variable is the sum of joint probabilities over the remaining components. For any m $(1 \le m \le n - 1)$, we denote by $S_m$ a m-dimensional random variable whose components are chosen from S and denote by $S_{n-m}$ the complementary (n-m)-dimensional random variable with respect to S. Let $v$ denote an arbitrary n-dimensional binary vector that S may take as value. Let $u$ denote an arbitrary m-dimensional binary vector and $w$ denote an arbitrary (n-m)-dimensional binary vector that $S_m$ and $S_{n-m}$ may take as values respectively. The marginal (or absolute) probability distribution $p(u) = \sum_w p(u,w)$ can be computed easily according to the binary semantic network. Similarly, the conditional probability distribution $p(u|w) = \frac{p(v)}{p(w)}$, where $p(w)$ has to be positive, can also be computed easily. Thus, we may compute joint generalized truth value (or probability) of any subset $S_m$ of S and the conditional probability of $S_m$ given the complement $S_{n-m}$ from the joint probabilities of S. This is an extension of [1] which deals with individual propositions. Moreover, we can derive the Bayes formulas easily for $S_m$ and $S_{n-m}$: $p(u) = \sum_w p(u|w).p(w)$, $p(w|u) = \frac{p(w).p(u|w)}{p(u)}$.

We may also compute joint probabilities (or permissible probabilistic interpretation vector) from marginal and conditional probabilities, provided that they are known. Here, we have a slight variation of Nilsson's probabilistic inference scheme which is described as follows. We are given a set S of propositions with known joint probabilities from which all marginal and conditional probabilities of S may be computed. For a new proposition s, we like to know its marginal probability in terms of the joint probabilities of the set S', the union of S and {s}. Construct a binary semantic tree for S' by appending the tree for S. We simply assign permissible conditional probabilities of s given S to the appended binary semantic tree to obtain the joint probabilities of S'. The marginal probability of s can be readily computed. Thus, we have an efficient approach to probabilistic logic.

## 4. CONCLUSION

We have presented some ideas of how to extend the probabilistic logic to an evidential one in the framework of Dempster-Shafer theory. The probabilistic interpretation of the probabilistic logic provides the setting of a relaxation scheme which extends some relaxation labeling schemes of computer vision. We maycan view probabilistic logic as a labeling problem which labels logical propositions by 0 and 1. Starting with an initial marginal and conditional probability assignments, the process will converge to a stable and deterministic labeling of the set S of logical propositions.